\documentclass[10pt,twocolumn,letterpaper]{article}

\usepackage{cvpr}              % To produce the CAMERA-READY version
\usepackage[dvipsnames]{xcolor}

\usepackage{graphicx}
\usepackage{amsmath}
\usepackage{amssymb}
\usepackage{booktabs}
\usepackage[dvipsnames]{xcolor}
\usepackage{multirow}
\usepackage{tikz}
\usepackage{pifont}
\usepackage{color, colortbl}
\definecolor{Gray}{gray}{0.9}

\definecolor{cvprblue}{rgb}{0.21,0.49,0.74}
\usepackage[pagebackref,breaklinks,colorlinks,citecolor=cvprblue]{hyperref}
\usepackage{tcolorbox}

\def\paperID{7924} % *** Enter the Paper ID here
\def\confName{CVPR}
\def\confYear{2024}
\definecolor{mypink}{RGB}{239,43,159}

\title{RLHF-V: Towards Trustworthy MLLMs via Behavior Alignment from \\ Fine-grained Correctional Human Feedback}

\author{
Tianyu Yu$^{1}$ \ \  
Yuan Yao$^2\thanks{Corresponding authors}$ \ \ \
Haoye Zhang$^1$ \ \
Taiwen He$^{1}$ \ \ 
Yifeng Han$^{1}$ \ \ \\  
Ganqu Cui$^1$   \ \
Jinyi Hu$^1$ \ \
Zhiyuan Liu$^{1*}$ \ \ 
Hai-Tao Zheng$^{1*}$ \ \ 
Maosong Sun$^1$ \ \ 
Tat-Seng Chua$^2$
\\[0.5em]
$^1$Tsinghua University \ \ \
$^2$National University of Singapore \ \ \ 
\ \ \ \\
{\small \texttt{yiranytianyu@gmail.com} \quad \texttt{yaoyuanthu@gmail.com}} \\
\\
\large{\color{mypink} \textbf{ \url{https://rlhf-v.github.io}}}
}

\begin{document}
\maketitle

\begin{abstract}
Multimodal Large Language Models (MLLMs) have recently demonstrated impressive capabilities in multimodal understanding, reasoning, and interaction. However, existing MLLMs prevalently suffer from serious hallucination problems, generating text that is not factually grounded in associated images. The problem makes existing MLLMs untrustworthy and thus impractical in real-world (especially high-stakes) applications. To address the challenge, we present RLHF-V, which enhances MLLM trustworthiness via behavior alignment from fine-grained correctional human feedback. Specifically, RLHF-V collects human preference in the form of segment-level corrections on hallucinations, and performs dense direct preference optimization over the human feedback. Comprehensive experiments on five benchmarks in both automatic and human evaluation show that, RLHF-V can enable substantially more trustworthy MLLM behaviors with promising data and computation efficiency. Remarkably, using 1.4k annotated data samples, RLHF-V significantly reduces the hallucination rate of the base MLLM by 34.8\%, outperforming the concurrent LLaVA-RLHF trained on 10k annotated data. The final model achieves state-of-the-art performance in trustworthiness among open-source MLLMs, and shows better robustness than GPT-4V in preventing hallucinations aroused from over-generalization.
\end{abstract}

\section{Introduction}

The recent success of Multimodal Large Language Models (MLLMs) marks a significant milestone in AI research~\cite{GPT4V,alayrac2022flamingo,li2023blip,liu2023visual,zhu2023minigpt,bai2023qwen,dai2023instructblip,wang2023cogvlm,driess2023palm}. By connecting visual signals and Large Language Models (LLMs), MLLMs show unprecedented capabilities in multimodal understanding, reasoning, and interaction~\cite{GPT4V,yang2023dawn,lu2023mathvista}. The models are typically pre-trained on large-scale image-text data to learn the foundational multimodal knowledge and capabilities ~\cite{alayrac2022flamingo,driess2023palm,li2023blip,bai2023qwen}. To steer the model behavior, most MLLMs are further fine-tuned with instruction tuning (also known as supervised fine-tuning), which supervises models to clone the behavior from demonstration data, enabling MLLMs to engage users in various open-world tasks~\cite{liu2023visual,dai2023instructblip,liu2023aligning,bai2023qwen,yu2023reformulating}.

However, current MLLM behaviors are not well aligned with human preferences. A glaring issue is their tendency to produce \textit{hallucinations} --- responses that are not factually grounded in the associated images~\cite{GPT4V,sun2023aligning,liu2023aligning,li2023evaluating}. This typically includes descriptions of non-existing visual contents and errors in descriptions. As shown in Figure~\ref{fig:overview}, current MLLMs can hallucinate about objects, attributes, numbers, positions, actions, etc. Quantitatively, our human evaluation shows that the problem is prevalent among state-of-the-art MLLMs, where even the most advanced GPT-4V~\cite{GPT4V} contains obvious hallucinations in 45.9\% responses. The problem makes existing MLLMs untrustworthy and thus impractical in real-world (especially high-stakes) applications, such as guiding visually impaired individuals~\cite{GPT4V} or autonomous driving systems~\cite{wen2023road}.

{
\definecolor{red}{RGB}{239, 69, 31}
\definecolor{green}{RGB}{112, 173, 71}
\begin{figure*}[t]
    \centering
    \includegraphics[width=\textwidth]{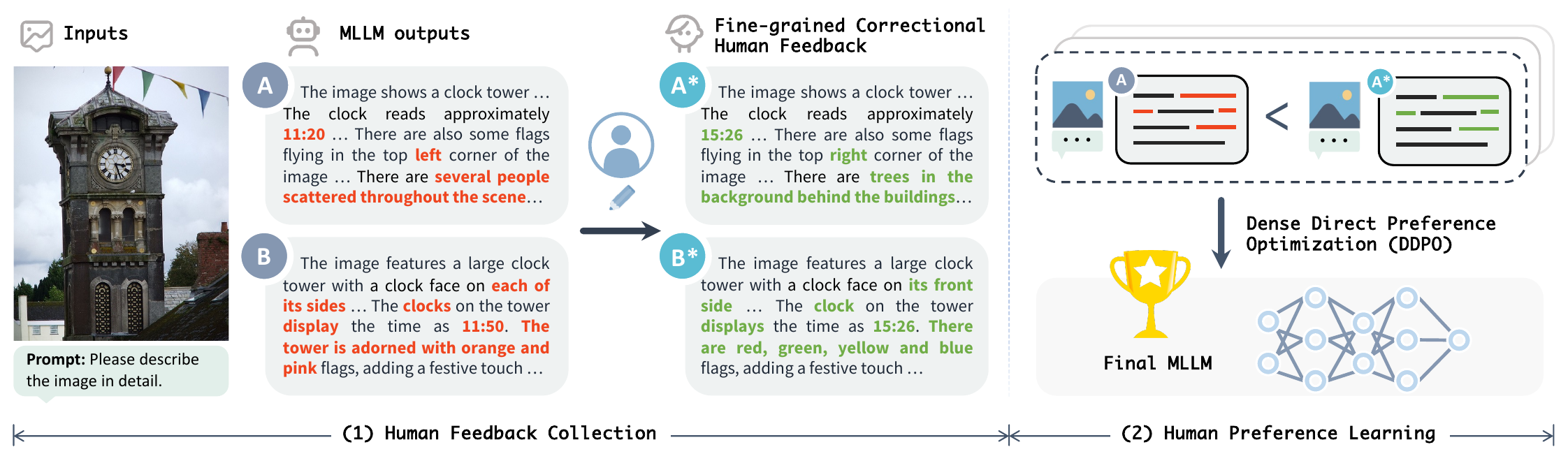}
    \caption{The RLHF-V framework for MLLM behavior alignment from human feedback. (1) Given the input image and prompt, we obtain outputs from MLLMs and collect human feedback in the form of fine-grained segment-level {\color{green}\textbf{corrections}} on {\color{red}\textbf{hallucinations}}. (2) During human preference learning, we perform dense direct preference optimization over the fine-grained correctional human feedback.}
    %\vspace{-5pt}
    \label{fig:overview}
\end{figure*}
}

We argue that the problem arises from the lack of positive/negative human feedback in instruction-tuned models, making it challenging to learn the precise behavior boundaries to exclude hallucination. To address the problem, we propose RLHF-V, a novel framework that aligns MLLM behavior by learning from human feedback. A straightforward way is to employ the traditional Reinforcement Learning from Human Feedback (RLHF) method in state-of-the-art LLMs~\cite{touvron2023llama,openai2023gpt4}, which involves human annotators ranking model responses, and utilizing a reward model to guide the policy LLM learning. However, this approach is fraught with two key challenges: (1) \textit{Annotation ambiguity}. Helpful and engaging responses about rich image content are typically long and complex, making it usually non-obvious to decide which response is preferable. As shown in Figure \ref{fig:overview} (responses A and B), annotators usually face dilemmas when presenting responses with respective advantages and flaws. Besides, even if labeled with a clear preference, the optimal response remains unknown (e.g., the exact time of the clock). (2) \textit{Learning efficiency}. The coarse-grained ranking feedback makes it difficult to accurately allocate credit to the desirable behaviors. Considering the linguistic complexity and variance of responses, the desirable behavior often requires a large amount of labeled data to learn~\cite{cui2023ultrafeedback,sun2023aligning,ouyang2022training}. Moreover, misallocation of credit to the non-robust bias correlated with the data usually leads to reward hacking and behavior degeneration problems~\cite{bai2022training,touvron2023llama}. 

% (3) \textit{Method complexity}. Traditional RLHF methods require training multiple LLMs to serve as reward, policy and value models, and extensively sampling the policy LLMs during the training loop, making the procedure complex and computationally expensive.

RLHF-V addresses these challenges by introducing two key innovations: (1) At the data level, we propose to collect human feedback in the form of fine-grained segment-level corrections. As shown in Figure~\ref{fig:overview}, we ask human annotators to directly correct the hallucinated segments from model responses, providing a clear, dense, and fine-grained human preference, as well as optimal responses. This strategy also avoids linguistic variance and non-robust bias, ensuring that the feedback is accurately allocated to the desirable behaviors, thereby enhancing learning efficiency and preventing reward hacking problems. (2) At the method level, we propose dense direct preference optimization (DDPO), a new variant of DPO~\cite{rafailov2023direct} that addresses the traditional RLHF objective in an equivalent simple and efficient supervised fashion. DDPO directly optimizes the policy model against dense and fine-grained segment-level preference, where the hallucinated segments receive stronger feedback to be factually grounded. 

% In addition, we also augment DDPO with supervised fine-tuning using high-quality instruction data generated by GPT-4V, which helps learning non-factual behaviors, such as response templates and styles.

Comprehensive experiments on five benchmarks show that, RLHF-V can substantially enhance the trustworthiness of MLLMs with promising data and computation efficiency. Using 1.4k preference data, RLHF-V significantly reduces the object hallucination rate of the base MLLM by 34.8\%, surpassing the concurrent LLaVA-RLHF~\cite{sun2023aligning} trained on 10k preference data. We also show that RLHF-V achieves better robustness than the strong GPT-4V~\cite{GPT4V} in preventing hallucinations aroused from over-generalization.

The contribution of this work can be summarized as threefold: (1) We present RLHF-V, a novel framework that aligns MLLM behavior through fine-grained correctional human feedback. (2) We collect high-quality human preference data to provide human-aligned learning signals for MLLMs. (3) We conduct comprehensive experiments to demonstrate the effectiveness of the proposed framework, achieving state-of-the-art performance in trustworthiness among open-source MLLMs. All the code, data, and model weights are open-sourced at \url{https://github.com/RLHF-V/RLHF-V}.

\section{Human Preference Collection}
% 怎么标注的，标什么数据，什么类型错误要标
\label{sec:data collection}
The goal of human preference data is to distinguish human-preferred high-quality responses from inferior ones, providing human-aligned learning signals to steer the MLLM behaviors. We first provide an analysis of underlying factors of human preference data, based on which we motivate the human preference collection procedure of RLHF-V.

\smallskip
\textbf{Human Preference Data: Underlying Factors and Challenges.} Given the input $x$ (including the image and the prompt), denote the difference between a preferred output $y_w$ and an inferior output $y_l$ as $Y$. The difference $Y$ can be essentially decomposed into three factors: 
\begin{equation}
    Y= Y_p + Y_s + Y_n,  
\end{equation}
where $Y_p$ is the truly preferred behavior such as being trustworthy and helpful, $Y_s$ denotes the shallow non-robust bias correlated with the data but unrelated to human judgment (e.g., $y_w$ contains more usage of specific words), and $Y_n$ is the random noise factor denoting the linguistic variance of natural language (e.g., different ways of expressing the same meaning). $Y_p$ is the factor we want to learn from the difference $Y$, while fitting to $Y_s$ can lead to reward hacking problems and thus should be avoided. The linguistic variance $Y_n$ does not bias the preference learning but makes the learning more difficult, demanding more labeled data to learn to the preferred factor $Y_p$, and thus should also be avoided if possible.

The common RLHF practices in LLMs collect human preference $Y$ in the form of ranking labels, indicating the overall relative quality of responses~\cite{touvron2023llama,ouyang2022training,openai2023gpt4}. According to the above analysis, the practice faces several key challenges: (1) \textit{Annotation ambiguity.} It can be non-obvious to annotate which response is superior using an overall ranking label due to the fine-grained nature of $Y_p$, especially for complex responses. As shown in Figure~\ref{fig:overview}, annotators usually cannot agree on assigning an overall ranking to different responses with respective advantages and flaws. We observe the issue leads to unsatisfactory annotation quality of existing RLHF data. Moreover, even if labeled with a clear preference, the optimal responses for the questions typically remain unknown. (2) \textit{Learning efficiency.} During reinforcement learning, it can be challenging and data-demanding to precisely allocate the sparse and coarse-grained credit from $Y$ through the linguistic variance $Y_n$ to the preferred behavior $Y_p$. Misallocation to the non-robust bias factor $Y_s$ will lead models to collapse to exploit trivial rewards~\cite{bai2022training,touvron2023llama}.

\smallskip
\textbf{Fine-grained Correctional Human Preference Collection.} To address the challenges, we propose to collect fine-grained human preferences in the form of segment-level corrections. As shown in Figure~\ref{fig:overview}, given a flawed output $y_l$ from MLLMs, we ask human annotators to directly correct the hallucinated segments, resulting in a factually optimal output $y_w$. The annotation simultaneously yields a segment-level incremental preference pair ($y_w$, $y_l$). The simple procedure effectively addresses the challenges: (1) The annotation of incremental correction in segments is clearer and more operable for human labelers. (2) The dense and fine-grained feedback is directly allocated to the preferred behavior $Y_p$, excluding the linguistic variance $Y_n$ and the non-robust bias $Y_s$, therefore improving learning efficiency and preventing reward hacking problems. In experiments, we find that the procedure greatly improves the annotation quality and data efficiency, enabling our model to surpass concurrent models trained on an order of magnitude more labeled preference data (see Section~\ref{sec:analysis}).

In practice, we obtain a total of 1.4k prompts as input from existing instruction tuning dataset~\cite{yu2023reformulating} and image description prompts generated by GPT-4, and get the responses from Muffin~\cite{yu2023reformulating} for human annotation. The responses after annotation contain 64.4 words and 2.65 corrected segments on average. We observe that the corrections are diverse in hallucination types, including objects (41.2\%), positions (20.3\%), numbers (16.5\%), attributes (10.0\%), actions (5.3\%) and miscellaneous types (6.8\%).

\section{Method}

We introduce the RLHF-V approach that learns the fine-grained correctional human feedback by dense direct preference optimization. In addition, we also mitigate existing sources of hallucination in MLLM training by addressing the vision-language mismatch problem.

% \subsection{Direct Fine-grained Preference Optimization}
% 公式过程
% \label{sec:DDPO}

\subsection{Dense Direct Preference Optimization}

To leverage the dense and fine-grained human feedback, we present DDPO, a new variant of direct preference optimization~\cite{rafailov2023direct} for directly optimizing the MLLM policy against dense human preference. The prevalent RLHF approaches involve fitting a reward model on the preference data, and then training the critique, policy and value models to maximize the reward without deviating too far from the reference model~\cite{touvron2023llama,ouyang2022training,cui2023ultrafeedback}. This procedure requires training multiple LLMs with extensive sampling and training, which suffers from complex procedures and high computation cost.

Direct Preference Optimization (DPO)~\cite{rafailov2023direct} solves this reinforcement learning objective in a simpler equivalent supervised fashion. Here we briefly introduce the DPO method, and refer readers to the original paper for more details. The key observation of DPO is that the reward function $r(x, y)$ can be analytically expressed by its optimal policy model $\pi_* (y|x)$ and reference model $\pi_\text{ref} (y|x)$, and therefore we can directly optimize the policy model under proper forms on the preference data. Specifically, the reward model $r(x, y)$ can be represented as:

\begin{equation}
\small
    r(x, y) = \beta \log \frac{\pi_* (y|x)}{\pi_\text{ref} (y|x)} + \beta \log Z(x),
\end{equation}
where $\beta$ is a constant and $Z(x)$ is the partition function. Based on this observation, the policy model can be directly optimized on the human feedback data:
% \vspace{-2pt}
\begin{equation}\label{eq:reward_model}
\resizebox{.9\hsize}{!}{%
$\begin{aligned}
    \mathcal{L} &= -\mathbb{E}_{(x, y_w, y_l)}\bigl[\log \sigma(r(x, y_w)- r(x, y_l))\bigr] \\
    &=  -\mathbb{E}_{(x, y_w, y_l)}\bigl[\log \sigma(\beta\log \frac{\pi_* (y_w|x)}{\pi_\text{ref} (y_w|x)}- \beta\log \frac{\pi_* (y_l|x)}{\pi_\text{ref} (y_l|x)})\bigr],
\end{aligned}$%
}
\end{equation}
where the reference model $\pi_\text{ref} (y|x)$ is usually implemented by an instruction-tuned base model we want to improve, and is kept fixed during DPO training. Only the policy model $\pi_* (y|x)$ is updated. We note that DPO is more simple, efficient and stable in aligning MLLM behaviors compared with traditional RLHF approaches. 

Leveraging dense and fine-grained segment-level feedback essentially requires the model to evaluate the reward of segment-level actions. However, DPO is designed for learning preference in the form of overall response ranking labels. Specifically, the action score of DPO is given by the likelihood of the holistic response in practice, where different segments are equally treated:

\begin{equation}
\small
    \log \pi(y|x) = \sum \limits_{y_i\in y} \log p(y_i|x, y_{<i}),
\end{equation}
where $y_i$ is the $i$-th token of the response $y$. We argue that compared with unchanged segments $y_u$, corrected segments $y_c$ more directly reveal human judgment in hallucination, and thus should contribute more to the overall action evaluation. Therefore, we propose to score the response as a weighted aggregation of the fine-grained segments:\footnote{For denotation simplicity, without confusion we also use $y_u$ and $y_c$ to denote the set of tokens in unchanged and corrected segments respectively.}

\begin{equation}
\resizebox{.9\hsize}{!}{%
  $\log \pi(y|x) = \frac{1}{N} \bigl[\sum \limits_{y_i\in y_u} \log p(y_i|x, y_{<i}) + \gamma \sum \limits_{y_i \in y_c} \log p(y_i|x, y_{<i})\bigr],$%
}
\end{equation}
where $\gamma > 1$ is a weighting hyperprameter, and larger $\gamma$ means more contribution from the corrected segments. $N = |y_u| + \gamma |y_c|$ is a normalizing factor, preventing longer responses from getting higher scores. In this way, corrected segments are highlighted to receive stronger human preference feedback to be factually grounded. In experiments, we find that DDPO can better exploit the fine-grained human feedback, leading to more trustworthy responses.

\subsection{Mitigating Hallucination from VL Mismatch} 
DDPO reduces hallucination by learning from human feedback. From another cause-and-effect view, we examine the mainstream MLLM training paradigm, and identify sources of hallucinations in training MLLMs. Based on the observations, we motivate a more trustworthy training recipe.

In general, current MLLMs learn multimodal capabilities in a supervised learning paradigm, where the model outputs are supervised against the ground-truth text associated with the image. In such a paradigm, hallucinations can be introduced by mismatches between images and text data. In practice, the mismatch can come from: (1) low-quality text in pre-training and instruction tuning data, and (2) careless image augmentation during training. We specify the issues and solutions in the following.

\smallskip
\textbf{Addressing Low-quality Text Influence.} Current pre-training data of MLLMs are automatically crawled from the Web~\cite{schuhmann2022laion,kakaobrain2022coyo-700m,changpinyo2021conceptual}, which inevitably suffers from severe noise in the text even after extensive post-processing. Supervising MLLMs against such data is essentially teaching them to hallucinate (e.g., describing elements not present in the image, or producing inconsistent descriptions with the image). Similarly, most existing visual instruction tuning datasets are generated by ChatGPT/GPT-4 according to intermediate text annotations~\cite{liu2023visual,liu2023aligning,yu2023reformulating}, which inevitably introduces hallucination into instruction data. While it can be difficult to repair existing pre-training and instruction-tuning data, we find that the influence can be countered by simply post-training MLLMs on high-quality visual question-answering datasets. Intuitively, human-labeled datasets can provide accurate learning signals to calibrate model behaviors from hallucinations, and also enhance instruction-following capabilities. In our experiments, we find that simply fine-tuning the model on VQAv2~\cite{goyal2017making} can significantly reduce the hallucination rate (see Section~\ref{sec:analysis}).

\smallskip
\textbf{Mitigating Untrustworthy Image Augmentation.} The vision-language mismatch can also come from the image domain. Data augmentation is widely adopted to improve the data diversity and model robustness in various multimodal models~\cite{radford2021learning,li2023blip,dai2023instructblip,yu2023reformulating,wang2023image}. However, we note that such augmentation must be performed with care in training MLLMs. The key problem is that some image augmentation operations can significantly change the semantics of images, which may make the augmented image inconsistent with the associated text. For example, during augmentation, random cropping can make the objects mentioned in the text absent from the image. This can make the model describe non-existing objects, with wrong numbers, and in wrong positions. In our model training, we exclude image cropping in data augmentation, which improves the trustworthiness of MLLMs (see Section~\ref{sec:analysis}).

% Considering the issue, we carefully go through common image augmentation methods, and divide them into two categories according to the above criteria: (1) Trustworthy ones include $x$, $x$ and $x$. (2) Untrustworthy ones include image cropping and flipping. During training, we only adopt trustworthy data augmentation approaches to prevent hallucinations.

\section{Experiments}
In this section, we empirically investigate the effectiveness of RLHF-V in aligning MLLM behaviors. In addition to evaluating the trustworthiness and helpfulness of conversation, we also analyze the data efficiency and scalability as well as the robustness. We refer readers to the appendix for more details on benchmarks, baselines and results.

\begin{table*}
    \centering
    \resizebox{\linewidth}{!}{
    \begin{tabular}{l c cc cccc cc ccc}
    \toprule
      \multirow{2}{*}{\textbf{Model}}   & \multicolumn{2}{c}{\textbf{Object HalBench} $\downarrow$} & \multicolumn{4}{c}{\textbf{MHumanEval} $\downarrow$} & \multicolumn{2}{c}{\textbf{MMHal-Bench}} & \multicolumn{3}{c}{\textbf{LLaVA Bench}} & {\textbf{VQAv2}} \\

      \cmidrule(lr){2-3} \cmidrule(lr){4-7} \cmidrule(lr){8-9}      \cmidrule(lr){10-12} \cmidrule(lr){13-13}
      
      & Resp.  & Mention & Object & Position & Number & All \hspace{1.2mm} &  \hspace{1.4mm}Info. & Resp.\hspace{0.3mm}$\downarrow$ & Conv. & Detail & Comp. & testdev \\

    \midrule
      LLaVA~\cite{liu2023visual}        & 63.0 & 29.5 &  46.6   &  21.2  & 19.9 &  80.8 & 31.9 & 70.8  & 85.4  & 74.3 & 96.3  & - \\
    Muffin~\cite{yu2023reformulating}       & 50.5 & 24.5 & 33.6 & 16.4 & 26.0 & 74.7 & 33.4 & 68.8 & 89.3 & \textbf{79.7} & \underline{97.7} & - \\
      LRV~\cite{liu2023aligning}   & 32.3    & 22.3    & 43.2  & \underline{11.6}  & {19.2}  & 82.9  & 22.2 & 78.1  & 61.7 & 47.3 & 55.0 & - \\
    % Muffin-QA~\cite{yu2023reformulating}       & 27.4 & 14.9 & 33.6    & 17.1    & 17.8  & 65.1  & \underline{39.6} & 58.3 & \underline{94.9}  & 76.1 & 90.9   & \textbf{80.0} \\
    LLaVA-RLHF~\cite{sun2023aligning}    & 38.1 & 18.9 & 37.7    & 17.8 & 18.5 & 72.6  & \underline{39.9} & 65.6 & \textbf{93.8} & 74.3 & \textbf{111.4} & - \\
      InstructBLIP~\cite{dai2023instructblip}   & \underline{25.9} & \underline{14.3} & \underline{30.8} & {15.1} & 17.1 & 63.7 & 29.5 & \underline{64.4} & 83.2 & 67.6 & 90.6  & - \\
      Qwen-VL-Chat~\cite{bai2023qwen} & 43.8    & 20.0    & 34.9    & {16.4}   & \underline{15.8} & \underline{61.0}  & 38.5 & \textbf{52.1} & 81.9 & \underline{77.1} & 92.3 & \underline{79.5} \\
      % 0.48
      LLaVA 1.5~\cite{liu2023improved}        & 46.3    & 22.6    & \underline{30.8}    & 17.8    & 17.1    & \underline{61.0}    & 39.2 &  \textbf{52.1}  & 81.6 & 75.5 & 95.2 & \textbf{80.0} \\
      % CogVLM-Chat        & 73.6    & 87.4    & 70.0    & 85.0    & 80.0    & 40.0    & 2.51 & 0.49   & - & - & - & - & 55.00 \\
      \midrule
      % RLHF-V (800)   & 96.0 & 96.1 & 76.7 & 83.3  & 90.0 & 41.7  & 2.58 & 0.51  & 89.5 &  66.3 & 84.7 & 80.1 & \textit{79.85} \\
      % RLHF-V (1300 + QA)  & 92.8 & 95.3 & - & -  & - & -  & 2.47 & 0.53 (0.40) & 86.2 &  58.1 & 83.9 & 76.0 & \textit{79.85} \\
  % RLHF-V (1300 + QA w SFT)  & 86.2 & 92.4 & - & -  & - & -  & - & -  & 92.0 &  72.1 & 91.4 & 85.2 & - \\
      RLHF-V  & \textbf{12.2} & \textbf{7.5} & \textbf{21.9} & \textbf{7.5}  & \textbf{14.4} & \textbf{55.5}  & \textbf{40.0} & \textbf{52.1}  & \underline{93.1} &  {75.3} & 91.6 & \textbf{80.0} \\
      % RLHF-V  & \textbf{11.1} & \textbf{6.6} & \textbf{20.6} & \textbf{15.1}  & \textbf{9.6} & \textbf{57.5}  & 39.3 & \underline{54.2}  & \textbf{95.0} &  \underline{76.7} & 89.1 & \textbf{80.0} \\
      % $\Delta$ & -16.3 & -8.3 & -13.0 & -2.0  & -8.2 & -7.6  & -0.3 & -4.1  & +0.1 &  +0.6 & -1.8 & +0.0 \\
      % RLHF-V (QA)  & 92.1 & 93.4 & - & -  & - & -  & 2.67 & 0.52  & - &  - & - & - & - \\
      % RLHF-V (only expansion)  & 91.1 & 91.4 & - & -  & - & -  & - & -  & 99.0 &  71.9 & 90.4 & 87.4 & - \\
      % RLHF-V (GPT-4V SFT)   & 89.5 & 92.4 & 41.7/90.5 & 75.0/96.4  & 83.3/98.6 & 25.0/80.2  & 2.77 & 0.46  & 94.6 &  95.3 & 89.7 & 88.6 & - \\
      \midrule
      \rowcolor{Gray}
      GPT-4V~\cite{GPT4V}       & 13.6 & 7.3 & 22.6 & 12.3  & 11.0 & 45.9  &   47.6  & 31.3  & 96.0 &  102.5 &  106.7  & \hspace{1mm} 77.2* \\
      
   \bottomrule
    \end{tabular}
    }
    \caption{Main experimental results on hallucination. We report hallucination rates in different granularities, including response-level (Resp.) and mention-level (Mention), and response-level hallucination rates in different types. We also show scores on informativeness (Info.), multimodal conversation (Conv.), detailed description (Detail), and complex reasoning (Comp.). * denotes zero-shot results on VQAv2.\protect\footnotemark[2] The best and second best open-source results are shown in \textbf{bold} and \underline{underlined} respectively.}
    \label{tab:main results}
\end{table*}

\subsection{Experimental Settings}
We first introduce the experimental settings, including evaluation, baselines, and implementation details.

\smallskip
\textbf{Evaluation.} We evaluate the models from two perspectives, including trustworthiness reflecting the hallucination degree, and helpfulness reflecting the general interaction quality. Similar to~\cite{sun2023aligning}, we find binary classification evaluation (i.e., answering yes/no)~\cite{li2023evaluating,fu2023mme} cannot well reflect the MLLM behaviors in open-ended long-form interactions. We thus adopt benchmarks that directly evaluate the long-form responses, which are more closely related to the practical usage scenarios of MLLMs. For trustworthiness, we perform evaluation on three benchmarks: 

(1) \textbf{Object HalBench}~\cite{rohrbach2018object} is a widely adopted benchmark for assessing object hallucination in detailed image descriptions. It compares the objects in the model output with object labels exhaustively annotated for COCO images~\cite{lin2014microsoft} to detect object hallucination. To improve the evaluation stability, we augment the benchmark with 8 diverse prompts for detailed image descriptions. We report the response-level hallucination rate (i.e., the percentage of responses that have hallucinations), as well as the mention-level hallucination rate (i.e., the percentage of hallucinated object mentions among all object mentions).

(2) \textbf{MMHal-Bench}~\cite{sun2023aligning} evaluates hallucinations and response informativeness. It employs GPT-4 to compare model output with human response and several object labels to decide the scores. In experiments, we find that GPT-4 cannot reliably detect hallucinations due to the incompleteness of MMHal-Bench text annotations. We therefore only report the informativeness score from GPT-4, and assess response-level hallucination rate by human evaluation. 

(3) \textbf{MHumanEval.} The above evaluations are either limited to common object hallucination or dominated by short-form question answering (i.e., questions that can be sufficiently answered by a few words). To provide a more reliable and comprehensive evaluation over diverse hallucination types, we present MHumanEval benchmark, which covers both long-form image descriptions, and short-form questions. The benchmark contains 146 samples collected from Object HalBench (50) and MMHal-Bench (96). Given model responses, we ask human annotators to label the hallucinated segments and hallucination types of the segments, including objects, positions, numbers and others. We report the response-level hallucination rate on these types. 

For helpfulness, we adopt two benchmarks: (1) \textbf{LLaVA Bench}~\cite{liu2023visual} is a widely adopted benchmark for assessing multimodal conversation, detailed description and complex reasoning capabilities. It scores model output against reference response via GPT-4. (2) \textbf{VQAv2}~\cite{goyal2017making} is a popular dataset for short-form visual question answering.

\footnotetext[2]{Due to limited instruction-following capability, most MLLMs need to be specifically fine-tuned to produce short-form VQA answers, and therefore cannot achieve reasonable zero-shot performance on VQAv2.}

\begin{table*}
    \centering
    % \resizebox{\linewidth}{!}{
    \small
    \begin{tabular}{l ccc ccc ccc ccc c}
    \toprule
      \multirow{4}{*}{\textbf{Model}}   & \multicolumn{3}{c}{\textbf{Living Room}} & \multicolumn{3}{c}{\textbf{Kitchen}} & \multicolumn{3}{c}{\textbf{Bathroom}} & \multicolumn{3}{c}{\textbf{Street}} & \multirow{4}{*}{$\overline{\Delta}$}\\
\vspace{-0.9mm}
     & \multicolumn{3}{c}{\scriptsize{book, person, bed}} & \multicolumn{3}{c}{\scriptsize{bottle, bowl, cup}} & \multicolumn{3}{c}{\scriptsize{{toilet, sink, bottle}}} & \multicolumn{3}{c}{\scriptsize{person, car, motorcycle}}\\

     & \multicolumn{3}{c}{\scriptsize{chair, couch, remote}} & \multicolumn{3}{c}{\scriptsize{person, chair, knife}} & \multicolumn{3}{c}{{\scriptsize{toothbrush, person, cup}}} & \multicolumn{3}{c}{\scriptsize{traffic light, handbag, truck}}\\

      \cmidrule(lr){2-4} \cmidrule(lr){5-7} \cmidrule(lr){8-10}      \cmidrule(lr){11-13}
      
      & $\text{H}_\text{a}$  & $\text{H}_\text{s}$ & $\Delta$ & $\text{H}_\text{a}$  & $\text{H}_\text{s}$ & $\Delta$ & $\text{H}_\text{a}$  & $\text{H}_\text{s}$ & $\Delta$ & $\text{H}_\text{a}$  & $\text{H}_\text{s}$ & $\Delta$  \\

    \midrule
      LLaVA-1.5~\cite{liu2023improved}	&25.2	&41.8	& +16.6	&18.9	&23.9	&+5.0	&22.4	&30.4	&+8.0	&20.6	&28.0	&+7.4	&+9.2 \\
      LLaVA-RLHF~\cite{sun2023aligning}	&23.7	&34.5	&+10.8	&13.1	&17.4	&+4.3	&18.2	&19.5	&+1.4	&18.3	&22.7	&+4.4	&+5.2 \\
      QWEN-VL~\cite{bai2023qwen}	&24.5	&34.5	&+10.0	&16.4	&20.8	&+4.4	&21.6	&17.5	&\textbf{-4.1}	&22.5	&32.0	&+9.5	&+5.0 \\
      RLHF-V	& \textbf{5.5}	&\textbf{8.0}	&\textbf{+2.5}	&\textbf{3.8}	&\textbf{5.9}	& \textbf{+2.1}	& \textbf{4.1}	&\textbf{4.0}	&-0.1	&\textbf{2.3}	&\textbf{4.6}	&\textbf{+2.3}	&\textbf{+1.7} \\
      \midrule
      \rowcolor{Gray} GPT-4V~\cite{GPT4V}	&8.2	&19.4	&+11.2	&4.6	&5.7	&+1.1	&5.9	&13.3	&+7.5	&4.2	&4.6	&+0.4	&+5.0 \\

   \bottomrule
    \end{tabular}
    % }
    \caption{Experimental results of hallucination from over-generalization on Object HalBench. For each scene, we report the hallucination rate of the top 10 frequent objects on average on the full benchmark ($\text{H}_\text{a}$) and under the scene ($\text{H}_\text{s}$). Top 6 frequent objects are listed for each scene for brevity. $\Delta$: hallucination rate difference, $\overline{\Delta}$: average difference across the scenes.} 
    \label{tab:scene}
\end{table*}

\smallskip
\textbf{Baselines.} We compare our model with state-of-the-art baselines. (1) \textbf{General baselines.} We adopt Qwen-VL-Chat~\cite{bai2023qwen}, LLaVA~\cite{liu2023visual}, LLaVA 1.5~\cite{liu2023improved}, Muffin~\cite{yu2023reformulating}, and InstructBLIP~\cite{dai2023instructblip} as representative general baselines. These models are mostly pre-trained on large-scale multimodal data, and fine-tuned on high-quality instruction data, achieving strong performance across various multimodal tasks. (2) \textbf{Baselines tailored for hallucination problems.} LRV~\cite{liu2023aligning} is fine-tuned on 400k instruction data generated by GPT-4, and mitigates hallucination by limiting the response length. The concurrent LLaVA-RLHF~\cite{sun2023aligning} employs the strong 13B Vicuna v1.5~\cite{zheng2023judging} (fine-tuned from LLaMA-2~\cite{touvron2023llama}) as LLM backbone. It trains the reward model on 10k human-labeled preference data, and performs proximal policy optimization~\cite{schulman2017proximal} on 72k factually augmented data. (3) \textbf{Commercial Baseline.} We also include GPT-4V~\cite{GPT4V} as a strong reference, evaluating the gap between the open-source models and state-of-the-art commercial models.

\smallskip
\textbf{Implementation Details.} We implement the RLHF-V framework based on Muffin~\cite{yu2023reformulating}. The model uses BEiT-3~\cite{wang2023image} as the visual module, and 13B Vicuna v1.0~\cite{vicuna2023} (fine-tuned from LLaMA~\cite{touvron2023llama1}) as the LLM backbone. The hyperparameter $\beta$ is 0.5, and the weighting coefficient $\gamma$ is 5. We train the model with DDPO for 7 epochs, with image resolution 448, learning rate 5e-7 and batch size 32. The training of RLHF-V is computationally efficient, which takes less than 1 hour on 8 A100 GPUs in total.

\begin{figure}[t]
    \centering
    \includegraphics[width=\linewidth]{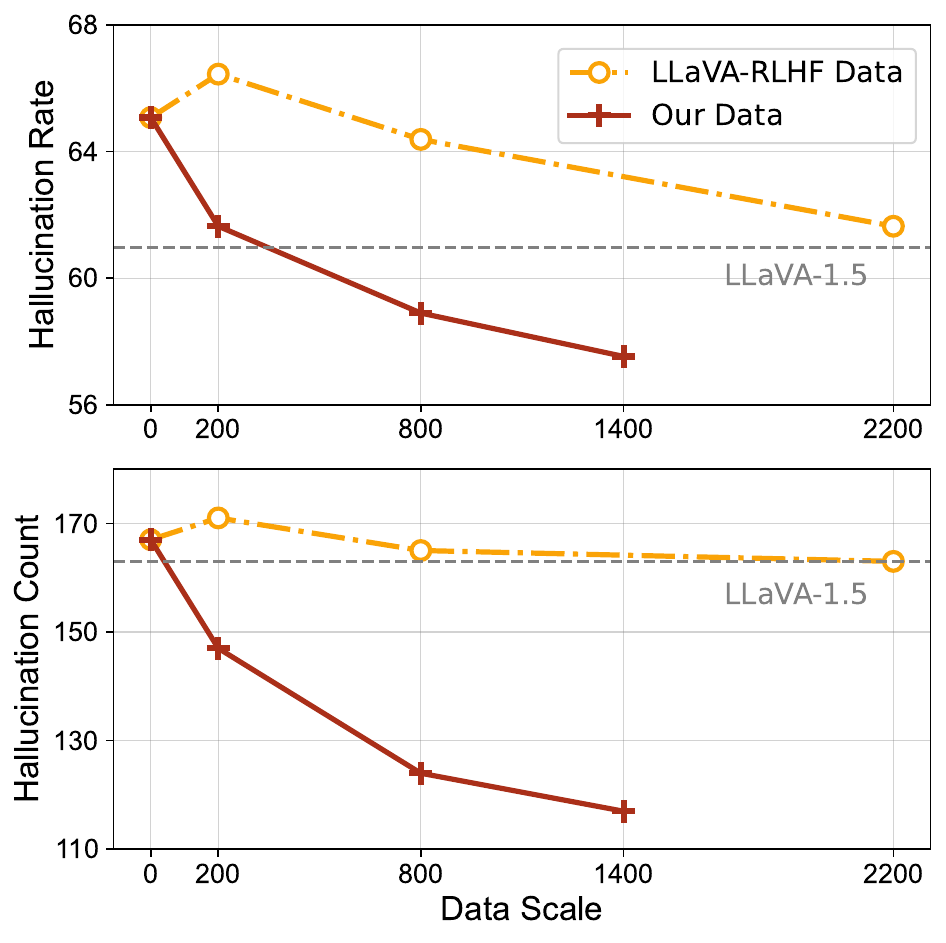}
    \caption{Hallucination rate and number on MHumanEval (all types) with respect to the amount of preference data. We report the results of different models trained on different RLHF data.}
    %\vspace{-5pt}
    \label{fig:data_scaling}
\end{figure}

\subsection{Main Results}

The main experimental results are reported in Table~\ref{tab:main results}, from which we observe that: (1) RLHF-V achieves state-of-the-art performance in trustworthiness among open-source models, outperforming strong general models and models tailored for hallucination. The framework significantly reduces the hallucination rate of the base model Muffin by 75.8\% relative points for common objects on Object HalBench, and by 34.8\% for overall objects on MHumanEval. The improvement is consistent in different granularities including response-level and mention-level hallucinations, and different hallucination types including objects, positions, and numbers. The reduction is more significant on the more challenging long-form answers on Object HalBench and MHumanEval. The results show that RLHF-V can effectively learn from fine-grained correctional human feedback to enable more trustworthy MLLM behaviors. (2) RLHF-V achieves promising performance in response helpfulness, where the results on MMHalBench, LLaVA Bench and VQAv2 are strong and comparable to the base model. This shows that RLHF-V can enhance the trustworthiness of MLLMs without sacrificing their helpfulness.

\subsection{Analysis}
\label{sec:analysis}
In this section, we conduct analyses on the framework considering the following research questions: (1) How does RLHF-V's performance scale with feedback data amount? (2)~What is the advantage of fine-grained correctional preference data over traditional overall ranking data? (3) Can RLHF-V's data and method be adopted to enhance the trustworthiness of other MLLMs? (4) How does human feedback alleviate hallucinations intuitively?

\smallskip
\textbf{Scaling feedback data leads to promising results.} We report the hallucination rate and numbers of hallucinated segments on MHumanEval under different amounts of feedback data in Figure~\ref{fig:data_scaling}. We observe that the hallucination rate and number of RLHF-V show a significant and rapid decrease as the data amount grows. This shows that fine-grained correctional human feedback provides effective and efficient learning signals for MLLM behavior alignment. Based on this tendency, we expect better performance can be achieved with an increasing amount of feedback data. We leave this for future work.

\smallskip
\textbf{Fine-grained correctional human feedback enables better learning efficiency.} To quantify the advantage of fine-grained correctional human feedback, we replace our data with the 2.2k human preference data on hallucination from LLaVA-RLHF, which gives overall ranking labels following common RLHF practices. From the experimental results in Figure~\ref{fig:data_scaling}, we observe that model equipped with our data shows a more significant and rapid reduction in hallucination rate and number. Notably, using only 200 preference data, our model achieves comparable hallucination rate to the model that uses an order of magnitude more labeled data from LLaVA-RLHF. The superior data efficiency is due to (1) better data quality since label ambiguity is minimized, and (2) more direct feedback on hallucinated segments, excluding non-robust bias and linguistic variance. 

\smallskip
\textbf{RLHF-V generalizes to enhance other MLLMs.} To investigate the generalization capability of the framework, we adopt RLHF-V's data and approach to align the behavior of LLaVA~\cite{liu2023visual}, a representative and widely used MLLM. Experimental results show that RLHF-V effectively reduces the hallucination count of LLaVA by 13.8 relative points, as well as the hallucination rate by 5.9 relative points. We also apply RLHF-V to stronger base models and build the OmniLMM-12B~\cite{omnilmm} which achieves new SoTA results on multiple hallucination benchmarks. For example, OmniLMM-12B exhibits only 4.5\% mention-level hallucination on the Object HalBench. Moreover, OmniLMM-12B also shows leading performance among comparable-sized models on multiple benchmarks (1637 on MME-Perception~\cite{fu2023mme}, 71.1 on SeedBench-I~\cite{li2023seed}). The results demonstrate that RLHF-V is applicable across different MLLMs to improve trustworthiness.

\begin{table}[t]
\small
\resizebox{\linewidth}{!}{
    \centering
    \begin{tabular}{lccc cc cccc cc}
    \toprule
          \multirow{2}{*}{\textbf{Model}} & \multicolumn{4}{c}{\textbf{MHumanEval}\hspace{0.5mm}$\downarrow$} & \textbf{MHB}\hspace{0.05mm}$\downarrow$ & \textbf{VQAv2}\\
      \cmidrule(lr){2-5}  \cmidrule(lr){6-6} \cmidrule(lr){7-7}  
          & Obj. & Pos. & Num. & All  & Resp. & testdev \\
          \midrule
          Muffin~\cite{yu2023reformulating} & 33.6 & 16.4 & 26.0 & 74.7 & 68.8
          & - \\ 
    \midrule
      % RLHF-V  & \textbf{12.2} & \textbf{7.5} & \textbf{20.6} & \textbf{9.6}  & \textbf{13.0} & \textbf{55.5}  & - & \underline{52.1}  & \underline{93.1} &  \underline{75.3} & 91.6 & \textbf{80.0} \\
        RLHF-V  & 21.9 & \textbf{7.5} & 14.4 & \textbf{55.5} & \textbf{52.1} & \textbf{80.0} \\
        \hspace{0.35em} w/ vanilla DPO  & 21.9 & 11.6 & \textbf{11.6} & 57.5 & 54.2 & \textbf{80.0} \\
        \hspace{0.35em} w/ IT-VQA only  & 34.3 & 17.1 & 17.1 & 65.1  & 58.3 & \textbf{80.0}\\
        % \hspace{0.35em} w/ IT-HF only & - & - & - & - & - & - \\
        \hspace{0.35em} w/ untrust aug.   & \textbf{18.5} & 13.7 & 14.4 & 59.6 & 54.2 & 77.1 \\
        %  \hspace{0.35em} w/o trust aug. & 11.7 & 23.3 & 12.3 & 10.3 & 58.2 & 52.1 \\
    \bottomrule
    \end{tabular}
    }
    \caption{Ablation results on different components. MHB: MMHal-Bench, IT-VQA: instruction tuning on VQAv2, untrust aug.: untrustworthy data augmentation.}
    \label{tab:ablation}
\end{table}

\smallskip
\textbf{RLHF-V reduces hallucination from correlation and over-generalization.} LLMs possess rich world knowledge and strong generalization capabilities. Without proper positive/negative human feedback, MLLMs can over-generalize to produce highly correlated and plausible concepts, which leads to hallucinations. For example, a prevalent hallucination case observed across different MLLMs is claiming the presence of \textit{person} as long as they see an image of \textit{street}. To quantify the problem, we select a set of representative scenes $\{\textit{living room}, \textit{kitchen}, \textit{bathroom}, \textit{street}\}$. For each scene, we identify the corresponding images in COCO by lexically matching the captions with the scene name. Then we obtain the top 10 frequent objects in the scene from the COCO object annotations. We compare the response-level hallucination rate for these objects (1) on average across all test samples, and (2) on samples under the target scene. Models prone to over-generalization will expect a significant increase in the hallucination rate ($\Delta$).

From the experimental results in Table~\ref{tab:scene}, we observe that: (1) All models including GPT-4V show a substantial increase in the hallucination rate, which demonstrates the over-generalization hypothesis. (2) RLHF-V exhibits the smallest change in the hallucination rate, which is even more robust than GPT-4V. The reason for the robustness is that RLHF-V provides crucial positive/negative fine-grained correctional human feedback for MLLMs, which helps to learn clear behavior boundaries between reasonable generalizations and over-generalizations. (3) RLHF-V achieves the lowest hallucination rates for these common objects both on average and especially under common scenes. This makes RLHF-V preferable in practical real-world applications.

\smallskip
\textbf{Ablation Study.} To investigate the contribution of each component, we perform an ablation study. From the experimental results in Table~\ref{tab:ablation}, we can observe that: (1) Learning human feedback with vanilla DPO leads to performance degrades, showing the advantage of DDPO in exploiting the fine-grained human preference. (2) Fine-tuning on VQAv2 leads to a significant reduction in hallucination rates compared with the base model. This reveals the value of traditional human-annotated datasets from a new perspective of hallucination mitigation. (3) Including untrustworthy data augmentation (i.e., image cropping) in training hurts the performance on both hallucination and VQAv2. This shows that careless data augmentation can be a double-edged sword in training MLLMs.

\definecolor{red}{RGB}{239,69,31}
% origin green 0,176,80
\definecolor{green}{RGB}{112, 173, 71} 
\definecolor{yellow}{RGB}{198,141,109}

\begin{figure*}[t]
    \centering
    \includegraphics[width=\textwidth]{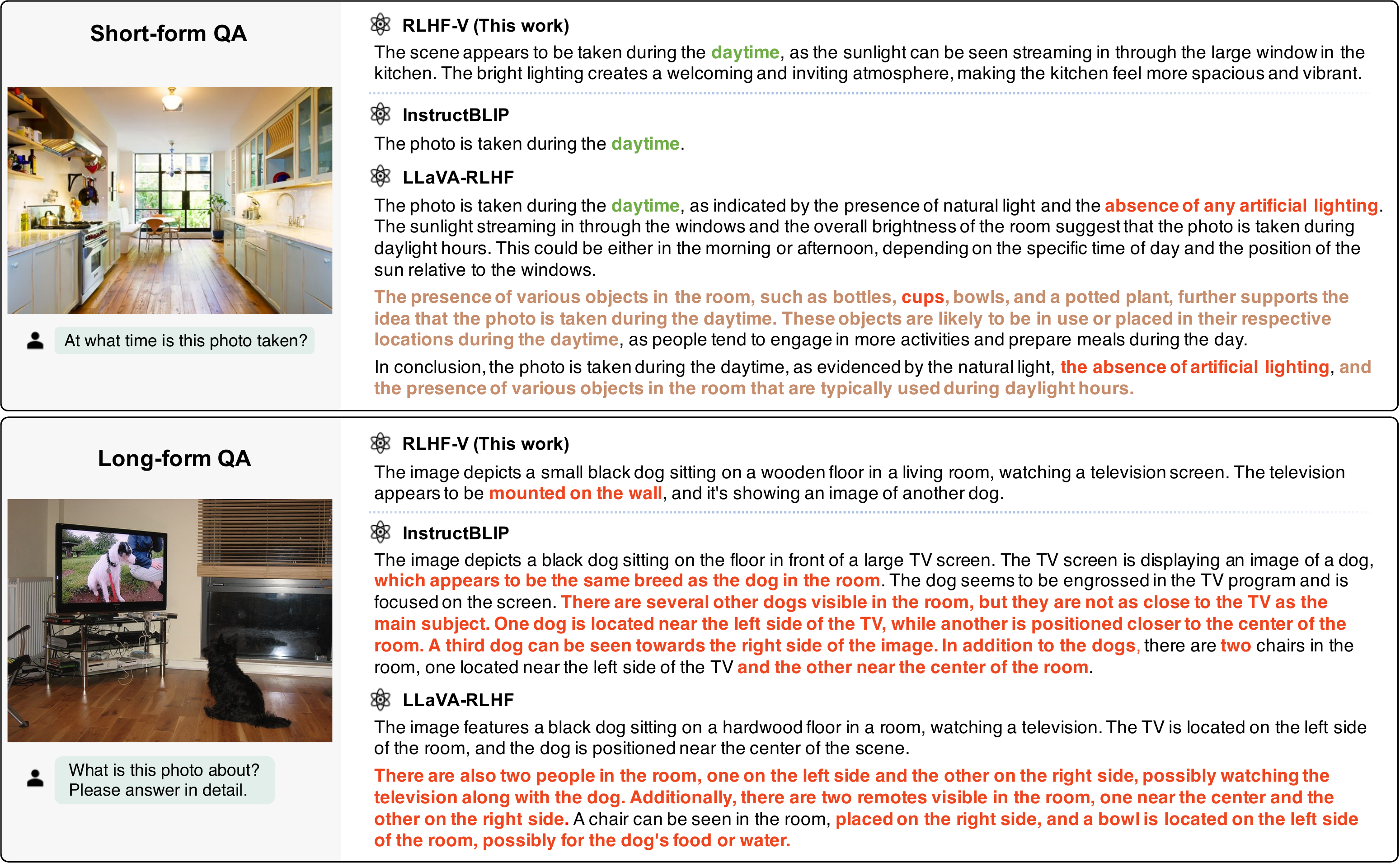}
    \caption{Qualitative results of different models on short-form QA and long-form QA. {\color{green}\textbf{Correct answers}}, {\color{yellow}\textbf{unreasonable extensions}} and {\color{red}\textbf{hallucinations}} are highlighted in color respectively.}
    %\vspace{-5pt}
    \label{fig:case}
\end{figure*}

\smallskip
\textbf{Case Study.} To provide an intuitive understanding and comparison of different models, we provide qualitative results in Figure~\ref{fig:case}. We show cases in two representative scenarios: (1) Short-form QA (i.e., questions that can be sufficiently answered in a few words). Our model typically maintains a good balance between helpfulness, engagement and clarity. In comparison, LLaVA-RLHF is usually far more engaging, introducing lengthy extensions however that can be less reasonable or relevant. (2) Long-form QA (i.e., questions that require long text to answer). We observe that MLLMs are significantly more prone to hallucinations in long-form QA, since it typically requires more comprehensive capabilities from multiple perspectives. For example, InstructBLIP and LLaVA-RLHF can confidently describe non-existing objects in a large proportion of their responses, whereas RLHF-V introduces significantly fewer hallucinations while delivering a comparable amount of effective information. We refer readers to the appendix for more qualitative results.

% \section{Discussion and Outlook}

% 可以用于语言模型
% 自动反馈
% alignment应该和基础能力匹配

\section{Related Work}
\textbf{Multimodal Large Language Models.} Recent trends in multimodal learning have witnessed the success of building MLLMs by connecting visual encoders with powerful LLMs~\cite{zhang2023llama,ye2023mplug,chen2022pali,huang2023language,li2023otter}. The current MLLM training paradigm typically involves two stages: (1) Pretraining. Models are pretrained on large-scale image-text pairs~\cite{bai2023qwen,wang2023cogvlm,dai2023instructblip,li2023blip,yu2023reformulating} or interleaved data~\cite{alayrac2022flamingo,huang2023language,awadalla2023openflamingo} to learn the semantic mapping between visual and text signals. (2) Instruction Tuning. To enable the model with instruction-following capability, MLLMs are further fine-tuned on visual instruction data, including collections of existing human-annotated datasets~\cite{dai2023instructblip,li2023m,liu2023improved}, and generated data from ChatGPT/GPT-4~\cite{liu2023visual,yu2023reformulating,liu2023aligning,li2023m}. Despite the success, current MLLMs suffer from serious hallucination problems~\cite{liu2023aligning,liu2023hallusionbench,sun2023aligning,li2023evaluating}. Notably, even after extensive efforts, GPT-4V has still been found to be prone to hallucinations, making basic factual errors confidently~\cite{GPT4V}. The problem undermines practical applications of MLLMs especially in high-stakes scenarios, which has recently drawn increasing attention from the community. 

\smallskip
\textbf{Behavior Alignment for LLMs.}
Aligning language agent behaviors with human preference has emerged as a promising research direction~\cite {leike2018scalable, kenton2021alignment}. Pivotal approaches in LLMs include instruction tuning (or supervised fine-tuning) and RLHF~\citep{stiennon2020learning, ouyang2022training}. While supervised fine-tuning is suitable for basic behavior alignment~\citep{taori2023stanford,ding2023enhancing}, due to the mismatch between likelihood maximization objective and human preference, it may introduce or amplify hallucination~\citep{ouyang2022training,openai2023gpt4}. Therefore, RLHF is widely accepted for further behavior and preference alignment~\citep{openai2023gpt4,bai2022constitutional,cui2023ultrafeedback}, where proximal policy optimization (PPO)~\citep{schulman2017proximal} is recognized as the major technique. Later adaptations attempt to stabilize the optimization process~\citep{rafailov2023direct} and enclose more fine-grained signals~\citep{lightman2023let,wu2023fine}.
However, RLHF has rarely been explored in MLLMs to align model behaviors with humans.

\smallskip
\textbf{Reducing Hallucination for MLLMs.} Some preliminary efforts have been made to alleviate hallucination problems in MLLMs. LRV~\cite{liu2023aligning} generates instruction data with negative responses, and mitigates hallucination by limiting the response length. However, limiting the response length does not essentially address the problem, and also undermines the response helpfulness. VIGC~\cite{wang2023vigc} iteratively refines the instruction data for better instruction tuning. Woodpecker~\cite{yin2023woodpecker} proposes to post-edit hallucinations by merging the output of MLLMs and a more accurate expert VQA model using GPT-3.5. The post-editing procedure involves external tools and LLMs much larger than the target MLLM online in multiple stages, which leads to high inference costs and delays. Gunjal \etal~\cite{gunjal2023detecting} distinguishes the inaccurate parts in responses via human annotation, and internally discourages the hallucinated parts by direct preference optimization. However, the positive behaviors for hallucinated parts are unknown, making the human feedback not complete enough to learn the behavior boundary. The concurrent LLaVA-RLHF~\cite{sun2023aligning} employs the traditional RLHF approach~\cite{ouyang2022training} on MLLMs, and augments the reward model with rich additional text descriptions. It is therefore similarly challenged with label ambiguity, learning efficiency, and complex training. In comparison, RLHF-V presents the first fine-grained correctional human feedback learning framework for behavior alignment, and systematically addresses different hallucination sources in training MLLMs, achieving strong performance in trustworthiness.

\section{Conclusion}
Hallucination is a critical problem preventing practical applications of MLLMs in real-world scenarios. In this work, we present RLHF-V, a novel framework that enhances the trustworthiness of MLLMs by behavior alignment from fine-grained correctional human feedback. Comprehensive experimental results show that our model achieves state-of-the-art performance in trustworthiness especially in challenging long-form responses while maintaining strong helpfulness.  In this work, we collect correctional feedback from human annotators. In future, with the progress of more trustworthy and capable MLLMs, we will explore collecting accurate preferences from MLLMs, which can facilitate large-scale preference learning for stronger behavior alignment. Besides, we note that the framework of RLHF-V can potentially also help reduce the hallucinations in LLMs, which we will explore in future.

\section*{Contributions}

The authors' contributions can be outlined as follows:

\begin{itemize}
    \item In initializing the project, Yuan Yao and Tianyu Yu design the framework to collect correctional human feedback. Tianyu Yu devise the DDPO algorithm. Zhiyuan Liu, Hai-Tao Zheng, Maosong Sun and Tat-Seng Chua offer invaluable guidance in project design. 

    \item In data collection, Taiwen He, Haoye Zhang, Tianyu Yu and Yuan Yao take charge of the annotation process to ensure the data quality.

    \item In model training and evaluation, Tianyu Yu implements the training framework. Tianyu Yu, Haoye Zhang and Yuan Yao design the evaluation framework. Tianyu Yu and Haoye Zhang implement the evaluation codebase.

    \item In paper writing, Yuan Yao and Tianyu Yu write the paper. Haoye Zhang, Taiwen He, Yifeng Han, Ganqu Cui, Zhiyuan Liu, Hai-Tao Zheng, Maosong Sun and Tat-Seng Chua offer suggestions to polish the writing.

    \item For public usability, Tianyu Yu, Yifeng Han, Jinyi Hu and Yuan Yao promote the open-source project.

    \item Throughout the project, Zhiyuan Liu, Hai-Tao Zheng, Maosong Sun and Tat-Seng Chua provide invaluable guidance and advice.

\end{itemize}

% We show that RLHF-V can generalize to different MLLMs, exhibit good data efficiency and scalability, and effectively cut off the over-generalization tendency.

{
    \small
    \bibliographystyle{ieeenat_fullname}
    \bibliography{main}
}

\clearpage
\appendix

\section{Zoom-in Study regarding GPT-4V}
We perform a zoom-in study of RLHF-V concerning GPT-4V to provide a better understanding of their behaviors. 

\subsection{Hallucination Patterns}

We conduct a comparative analysis of the responses generated by RLHF-V and GPT-4V, and have the following key observations:

(1) Compared with RLHF-V, GPT-4V tends to describe more details in the images and elaborate more on the interrelations among them. Quantitatively, we utilize ChatGPT to extract all the object mentions in the responses of GPT-4V, and find that the average number per response is 2.1 times larger than RLHF-V. We mainly attribute this to the higher resolution (7.6 times larger than RLHF-V)~\cite{GPT-4V_resolution} and the more powerful LLM backbone~\cite{GPT4V}. 

\begin{figure}
    \centering
    \includegraphics[width=\linewidth]{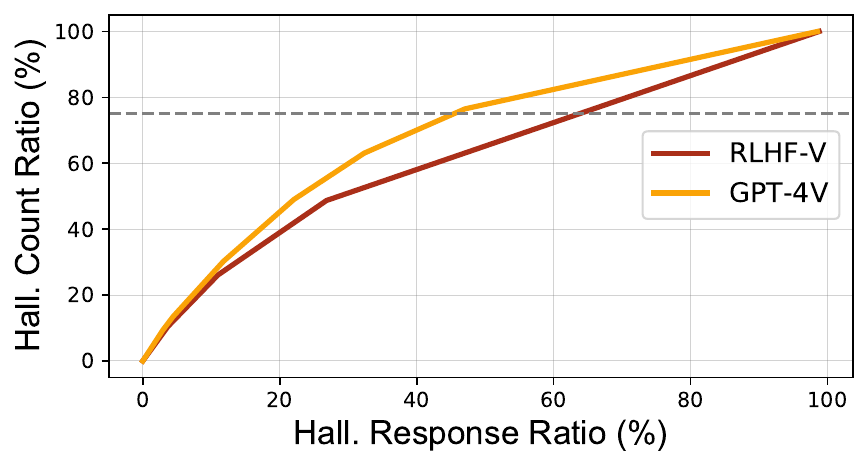}
    \caption{Distribution of hallucination segments over different responses. GPT-4V hallucinations are more concentrated on a smaller subset of the responses. Hall.: Hallucination.}
    \label{fig:hall_accumulation}
\end{figure}

(2) GPT-4V's hallucinations are more concentrated in some responses. In HumanEval, the hallucination rates of GPT-4V on \textit{Object} and \textit{Position} are comparable to RLHF-V. However, in the comprehensive \textit{ALL} metric, the hallucination rate is 17.3\% lower than RLHF-V. To better understand the reasons behind this phenomenon, we conduct a thorough analysis of the evaluation results. We observe that different types of hallucinations in GPT-4V are often concentrated in a small subset of responses, while contributing to hallucination rates across multiple subcategories. Quantitatively, we sort the responses of each model by the hallucination count in descending order, and plot the curve of hallucination count ratio vs hallucination response ratio. From the results in Figure~\ref{fig:hall_accumulation}, we can see that the top 45.6\% hallucinated responses of GPT-4V contribute to 75\% hallucination counts. In comparison, the top 64.6\% hallucinated responses of RLHF-V contribute to 75\% hallucinations. We refer readers to Section~\ref{sec:cases} for more qualitative results.

\subsection{Distillation against GPT-4V}

Upon observing GPT-4V's superior fine-grained image perception and text generation capabilities, an intuitive question is, will it be beneficial to distill GPT-4V capabilities through visual instruction tuning?  To this end, we collect 1.2k visual instruction data about long-form image descriptions from GPT-4V. We then use the response generated by GPT-4V to fine-tune our model. We observe that the average number of object mentions in the model response significantly increases by 1.8 times compared with the origin model. However, this can be a double-edged sword: as shown in Table~\ref{tab:gpt4_distill}, the hallucination rate significantly increases as well. 

The results are consistent with the hypothesis of~\cite{Hallucination_origin}: ``If we supervise the model against instruction data that far exceeds its own foundational capabilities, we are essentially teaching the model to hallucinate." Specifically, our model learns to produce more details and the interrelationship among them through distillation against GPT-4V, while the fundamental capabilities of the model are not enough for this demand. As a result, the hallucination problem is remarkably exacerbated. The results show that visual instruction data (or distillation target) is not the stronger the better, but rather should match the foundational capability of the model.

\section{Qualitative Results}
\label{sec:cases}

We provide more qualitative results in this section to facilitate a more intuitive understanding and comparison of different models. Based on the qualitative results, we have the following observations:

(1) RLHF-V typically exhibits less hallucination in both short-form QA and long-form QA scenarios, compared with open-source models such as LLaVA-RLHF and InstructBLIP, as shown in Figure \ref{fig:case_1}, \ref{fig:case_2}, \ref{fig:case_5}, and \ref{fig:case_5-2}. 

(2) GPT-4V is more descriptive regarding details in images as shown in Figure \ref{fig:case_2}, \ref{fig:case_5-2}, \ref{fig:case_3} and \ref{fig:case_4}. For example, in Figure \ref{fig:case_3}, GPT-4V mentions \textit{black dots} across each \textit{tile} while RLHF-V does not describe these details. 

(3) RLHF-V is more resistant to the over-generalization problem as shown in Figure \ref{fig:case_3} and Figure \ref{fig:case_4}. In Figure \ref{fig:case_3}, GPT-4V falsely mentions objects which are highly related to the scene while not shown in the image such as \textit{exhaust}, \textit{hood}, and \textit{bottle}.

% Figure \ref{fig:case_1} shows that RLHF-V can generate high-quality responses free of hallucinations when all baseline models including GPT-4V fail. Figure \ref{fig:case_2} shows that GPT-4V can generate more objects that are shown in the image compared to RLHF-V while keep being consistent with the image content.
% In Figure \ref{fig:case_3}, we can find even GPT-4V exhibits the over-generalization problem and tends to introduce objects that are highly related to kitchen while not shown in the image of kitchen such as \textit{range hood}, \textit{dish driving rack} and\textit{bottle}. On the other side, RLHF-V faithfully describe objects that are shown in the image.

\begin{table}[t]
\small
% \resizebox{\linewidth}{!}{
    \centering
    \begin{tabular}{l cccc c}
    \toprule
          \multirow{2}{*}{\textbf{Model}} & \multicolumn{4}{c}{\textbf{HumanEval}\hspace{0.5mm}$\downarrow$} & \textbf{MHB}\hspace{0.05mm}$\downarrow$\\
      \cmidrule(lr){2-5}  \cmidrule(lr){6-6}
          & Obj. & Pos. & Num. & All  & Resp.  \\
          \midrule
          Muffin~\cite{yu2023reformulating} & 33.6 & 16.4 & 26.0 & 74.7 & 68.8
           \\ 
    \midrule
        RLHF-V  & \textbf{21.9} & \textbf{7.5} & \textbf{14.4} & \textbf{55.5} & \textbf{52.1}  \\
        
        \hspace{0.35em} w/ GPT-4V distil.   & 45.2 & 10.3 & 20.6 & 75.3 & 62.5 \\
    \bottomrule
    \end{tabular}
    % }
    \caption{Experimental results of distillation against GPT-4V. MHB: MMHal-Bench, GPT-4V distil.: instruction-tune the model using responses generated by GPT-4V.}
    \label{tab:gpt4_distill}
\end{table}

\section{Implementation Details}

We provide more implementation details in this section for better reproducibility.
Benefiting from the high efficiency of training, we make all parameters trainable during the training process, which costs merely less than 1 hour on 8 A100 GPUs in total. We empirically find that adopting a longer warm-up (10\% training steps) can make the training more stable and consequently apply this setting for all experiments in this paper. As for data collection, besides the prompts obtained from~\cite{yu2023reformulating}, we also use image description prompts generated by GPT-4 during the annotation process which are listed in Table \ref{tab:gpt4_anno_prompt}.

\section{Evaluation Details}

We introduce more evaluation details, including baseline models and evaluation benchmarks.

\begin{table*}[h!]
\centering

\begin{minipage}{0.99\linewidth}\vspace{0mm}    \centering

\begin{tcolorbox} 
    % \centering
    \small
    \hspace{-6mm}

\begin{itemize}
    \item Identify and describe each object in the image in detail.
    \item Describe the key features of the image in great detail.
    \item What are the main elements in this image? Describe them thoroughly.
    \item Explain what's happening in the image with as much detail as possible.
    \item Detail the image's components with particular focus on each entity.
    \item Provide an intricate description of every entity in the image.
    \item What are the main objects or subjects in the image? Please describe them in detail.
    \item What is the setting or environment in which the image takes place?
    \item How do the elements in the image relate to each other in terms of positioning or composition?
    \item Explain the elements of the image with thorough attention to detail.
    \item Explain the image's various components in depth.
    \item What are the key features you observe in the image?
    \item Can you point out the details that make this image unique?
    \item Itemize the elements you identify in the image and describe them thoroughly.
    \item Convey the specifics of the image with meticulous attention to detail.
    \item Tell me what catches your eye in the image, and describe those elements in depth.
\end{itemize}

\end{tcolorbox}
    
\vspace{-2mm}
\caption{The list of instructions for detailed image description used in training.}
\label{tab:gpt4_anno_prompt}
\end{minipage}
\end{table*}

\begin{table*}[h!]
\centering

\begin{minipage}{0.99\linewidth}\vspace{0mm}    \centering

\begin{tcolorbox} 
    % \centering
    \small
    \hspace{-6mm}

\begin{itemize}
    \item Provide a thorough description of the given image.
    \item What is this photo about? Please answer in great detail.
    \item Provide a thorough description of the given picture.
    \item Explain the narrative or story that the image seems to convey, detailing each part that contributes to it.
    \item Please provide a detailed description of the image. Describe the visual elements, colors, shapes, textures, and any objects or people present along with the overall mood or atmosphere portrayed in the image.
    \item Please provide a detailed description of the image, including its visual elements, such as colors, shapes, textures, objects, and people.
    \item Provide an intricate description of the image, capturing its visual elements, including colors, shapes, textures, objects, and any people present.
    \item Compose a detailed account of the image, encompassing its visual characteristics, like colors, shapes, textures, objects, and any human subjects, by paying careful attention to the specifics.
\end{itemize}

\end{tcolorbox}

\vspace{-2mm}
\caption{The list of instructions for Object HalBench.}
\label{tab:obj_hall_bench}
\end{minipage}
\end{table*}

\begin{table*}[h!]
\centering

\begin{minipage}{0.99\linewidth}\vspace{0mm}    \centering

\begin{tcolorbox} 
    % \centering
    \small
    \hspace{-6mm}

You are an expert in image objects extraction according to a question answer pair. We asked an examiner to answer a question about a picture.

[Start of Question]

{\textless}image{\textgreater} \{question\}

[End of Question]

[Start of Examiner's Answer]

\{answer\}

[End of Examiner's Answer]

Assume that the answer is correct, please identify all visible objects that are directly shown in the image. Please following the instructions in below:

1. You should only mention objects that are explicitly mentioned in the examiner's answer.

2. You should only extract the object names without the attributes of the objects.

3. You should not include the properties of the object, like the color, material, etc. as part of the object name in your result.

4. Make your answer precise. Present the results in a JSON list format: [``object\_1'', ..., ``object\_n''].

5. You should return an empty JSON list () if no visible objects can be found.

\end{tcolorbox}
    
\vspace{-2mm}
\caption{The prompt we used to extract object mentions from image captions with ChatGPT.}
\label{tab:chair_extraction}
\end{minipage}
\end{table*}

\subsection{Baselines}

We compare with a series of state-of-the-art baselines:

\begin{itemize}
\item  \textbf{LLaVA}: LLaVA~\cite{liu2023visual} constructs 150K multimodal instructions based on the COCO dataset by asking GPT-4 to generate multi-turn dialogues for each image.
\item \textbf{Muffin}: Muffin~\cite{yu2023reformulating} propose to reformulate pre-trained vision-language models as the bridge toward large language models. The model is firstly pre-trained on 180M image-text pairs and then fine-tuned on their proposed UniMM-Chat instruction dataset consisting of 1.1M multimodal instructions.
\item \textbf{LRV}: LRV~\cite{liu2023aligning} is fine-tuned on 400K instruction data generated by GPT-4, and mitigates hallucination by limiting the response length.
\item \textbf{LLaVA-RLHF}: The concurrent LLaVA-RLHF employs the strong 13B Vicuna 1.5~\cite{zheng2023judging} (fine-tuned from LLaMA-2) as LLM backbone. It first trains the model with 122K instructions from VQAv2~\cite{goyal2017making}, A-OKVQA~\cite{AOKVQA} and Flickr30k~\cite{plummer2015flickr30k} to improve the foundational capabilities of the model. It then trains the reward model on 10K human-labeled preference data, and performs proximal policy optimization~\cite{schulman2017proximal} on 72K factually augmented data.
\item \textbf{InstructBLIP}: InstructBLIP~\cite{dai2023instructblip} constructs a multimodal instruction tuning dataset based on 26 public datasets by apply pre-defined templates to directly formulate these datasets into a unified format. They also devise a novel instruction-aware Q-Former and train the model on the proposed dataset.
\item \textbf{Qwen-VL-Chat}: Qwen-VL-Chat ~\cite{bai2023qwen} utilizes a large ViT with 1.9B parameters initialized from OpenCLIP's bigG~\cite{ilharco_gabriel_2021_5143773} as image encoder. It is pre-trained on 1.4B image-text pairs and fine-tuned on more than 50M high-quality multimodal instructions.
\item \textbf{LLaVA 1.5}: LLaVA 1.5~\cite{liu2023improved} also employs the strong 13B Vicuna 1.5~\cite{zheng2023judging} (fine-tuned from LLaMA-2) as LLM backbone. It is pre-trained on 558K selected image-text pairs and fine-tuned on 665K multimodal instructions with elaborately designed training strategies.

\end{itemize}

\subsection{Benchmarks}

We introduce additional details about the benchmarks we used for evaluation.

\begin{itemize}
\item  \textbf{Object HalBench}: Object HalBench~\cite{rohrbach2018object} is a widely adopted benchmark for assessing object hallucination in detailed image descriptions. To improve the evaluation stability, we augment the benchmark with 8 diverse prompts for detailed image descriptions during evaluation, where 4 instructions are adopted from~\cite{gunjal2023detecting} and the other 4 instructions are generated by GPT-4. We confirm that there is no overlap between the evaluation instructions and the training instructions. Detailed instructions are listed in Table \ref{tab:obj_hall_bench}. Following \cite{li2023evaluating}, we randomly sample 300 images from the validation set of COCO~\cite{lin2014microsoft} to form the evaluation image set. Regarding metrics, the response-level hallucination rate is the number of responses with object hallucinations divided by the number of responses that introduce COCO objects, while the mention-level hallucination rate is the number of falsely mentioned COCO objects in the generated responses divided by the total number of mentioned COCO objects. During evaluation, we first generate descriptions on images from the benchmark and then leverage ChatGPT to extract the mentioned objects in these responses which are further used to calculate the final scores following. Unlike  \cite{rohrbach2018object} which detects object mentions by exact-match, we find ChatGPT can perform the extraction with both better precision and recall and consequently apply this setting during evaluation. The full prompt we used to conduct such extraction is shown in Table \ref{tab:chair_extraction}.

\item  \textbf{MMHal-Bench}: MMHal-Bench~\cite{sun2023aligning} evaluates hallucinations and response informativeness. It consists of 96 images from the validation and test sets of OpenImages~\cite{open_images}. Each image in this benchmark is annotated with a brand new question and the image-question pairs cover 12 common object meta-categories from COCO.

\item \textbf{HumanEval}: The above evaluations are either limited to common object hallucination or dominated by short-form question answering (i.e., questions that can be sufficiently answered by a few words). To provide a more reliable and comprehensive evaluation over diverse hallucination types, we present HumanEval benchmark, which covers both long-form image descriptions, and short-form questions. The benchmark contains 146 samples collected from Object HalBench (50) and MMHal-Bench (96). Given model responses, we ask human annotators to label the hallucinated segments and hallucination types of the segments, including objects, positions, numbers and others. We report the response-level hallucination rate on these types.

\item \textbf{LLaVA Bench}: LLaVA Bench~\cite{liu2023visual}  is a widely adopted benchmark for assessing multimodal conversation, detailed description and complex reasoning capabilities. It consists of 30 image-question pairs for the aforementioned three capabilities respectively and scores model output against reference response via GPT-4.

\item \textbf{VQAv2}: VQAv2~\cite{goyal2017making} is a dataset for short-form visual question answering. The test-dev set of VQAv2 consists of 107K image-question pairs which covers a diverse range of capabilities.

\end{itemize}

\definecolor{red}{RGB}{239,69,31}
\definecolor{darkred}{RGB}{192,0,0}
\definecolor{green}{RGB}{112, 173, 71} 
\definecolor{yellow}{RGB}{198,141,109}

\begin{figure*}
    \centering
    \includegraphics[width=0.95\linewidth]{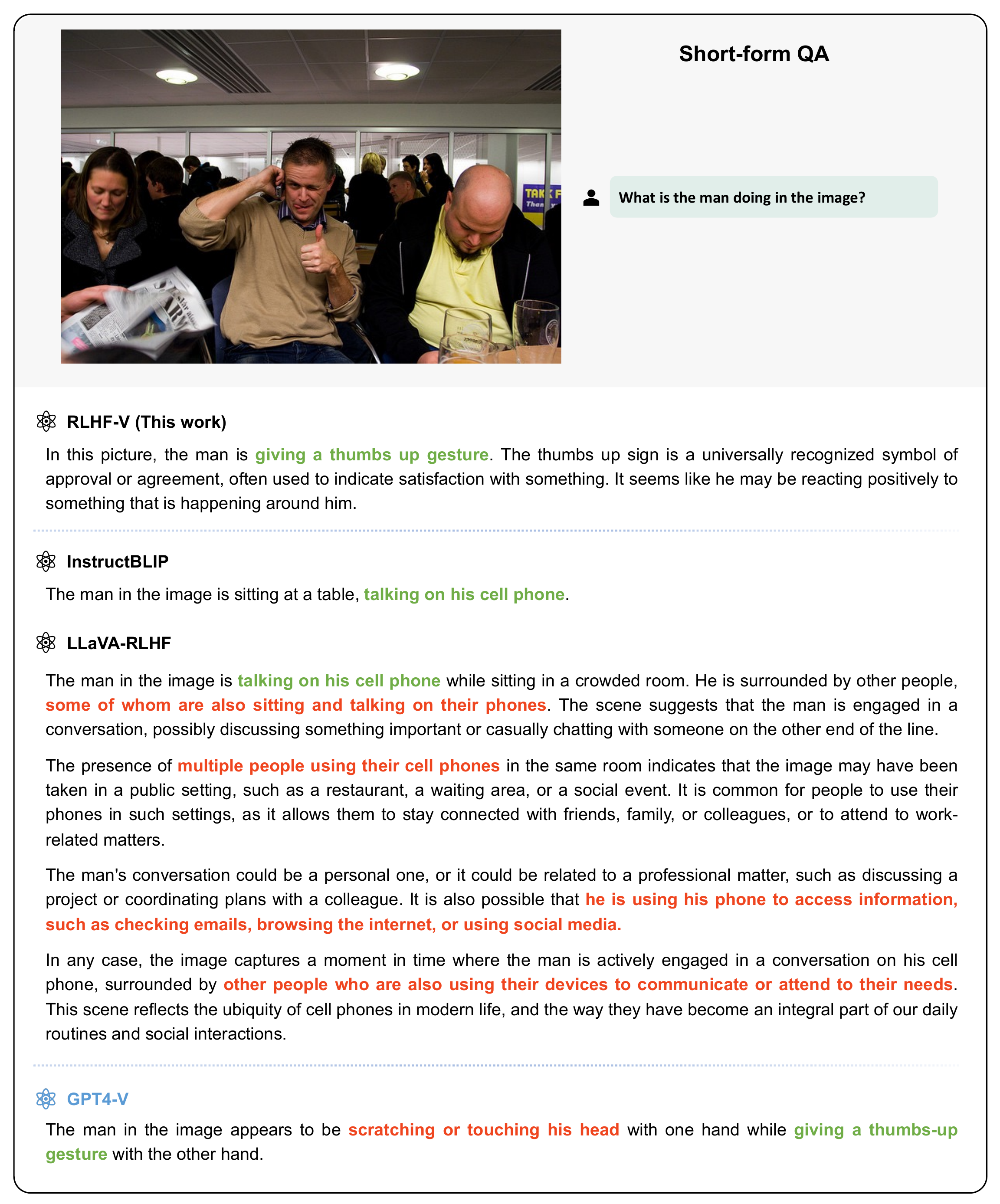}

    \caption{Qualitative results of different models. 
 {\color{green}\textbf{Correct answers}} and {\color{red}\textbf{hallucinations}} are highlighted in color respectively.}
    \label{fig:case_1}
\end{figure*}

\begin{figure*}
    \centering

    \includegraphics[width=0.95\linewidth]{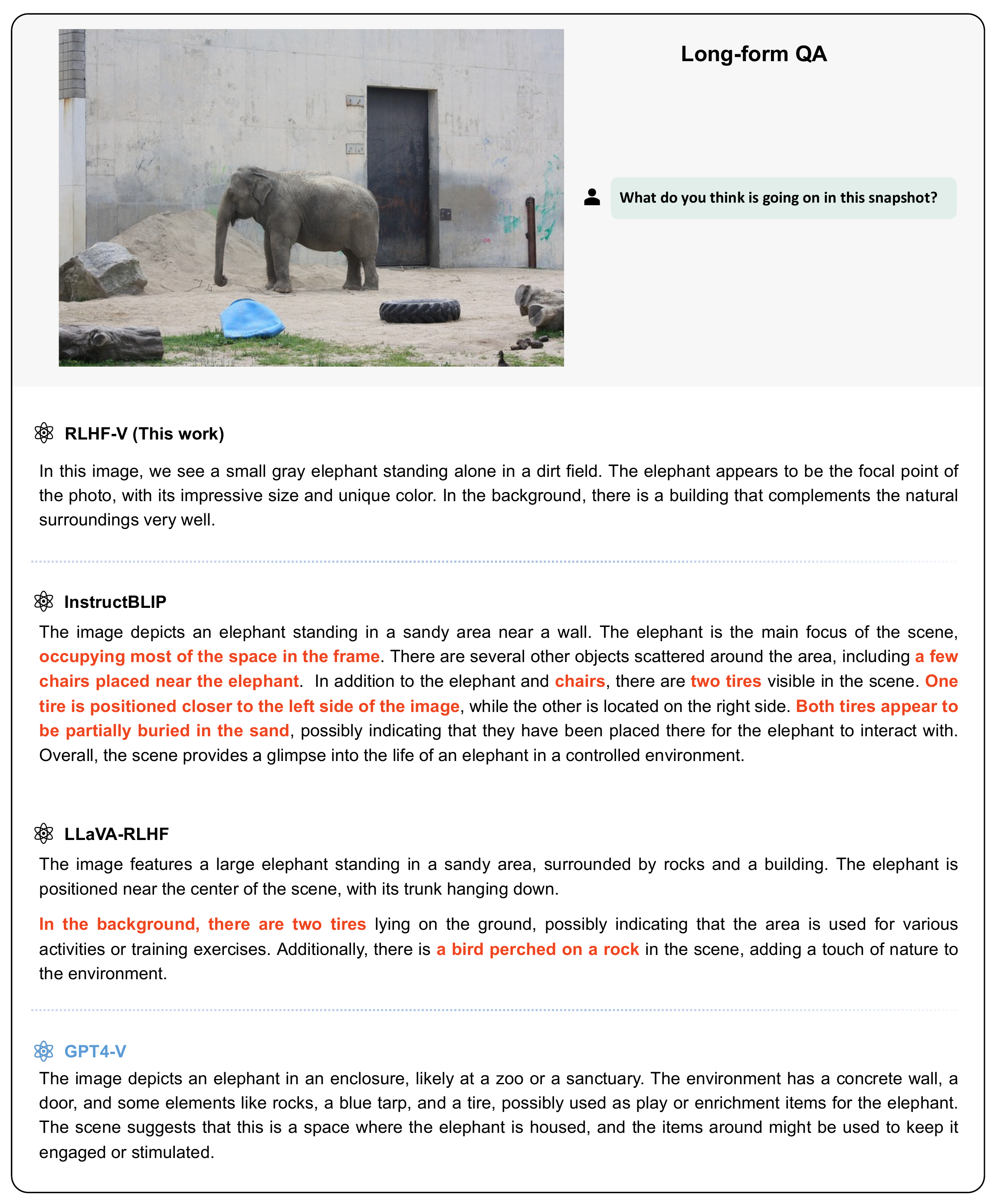}
    
    \caption{Qualitative results of different models. 
  {\color{red}\textbf{Hallucinations}} are highlighted in color.}
    \label{fig:case_2}
\end{figure*}

\begin{figure*}
    \centering

    \includegraphics[width=0.95\linewidth]{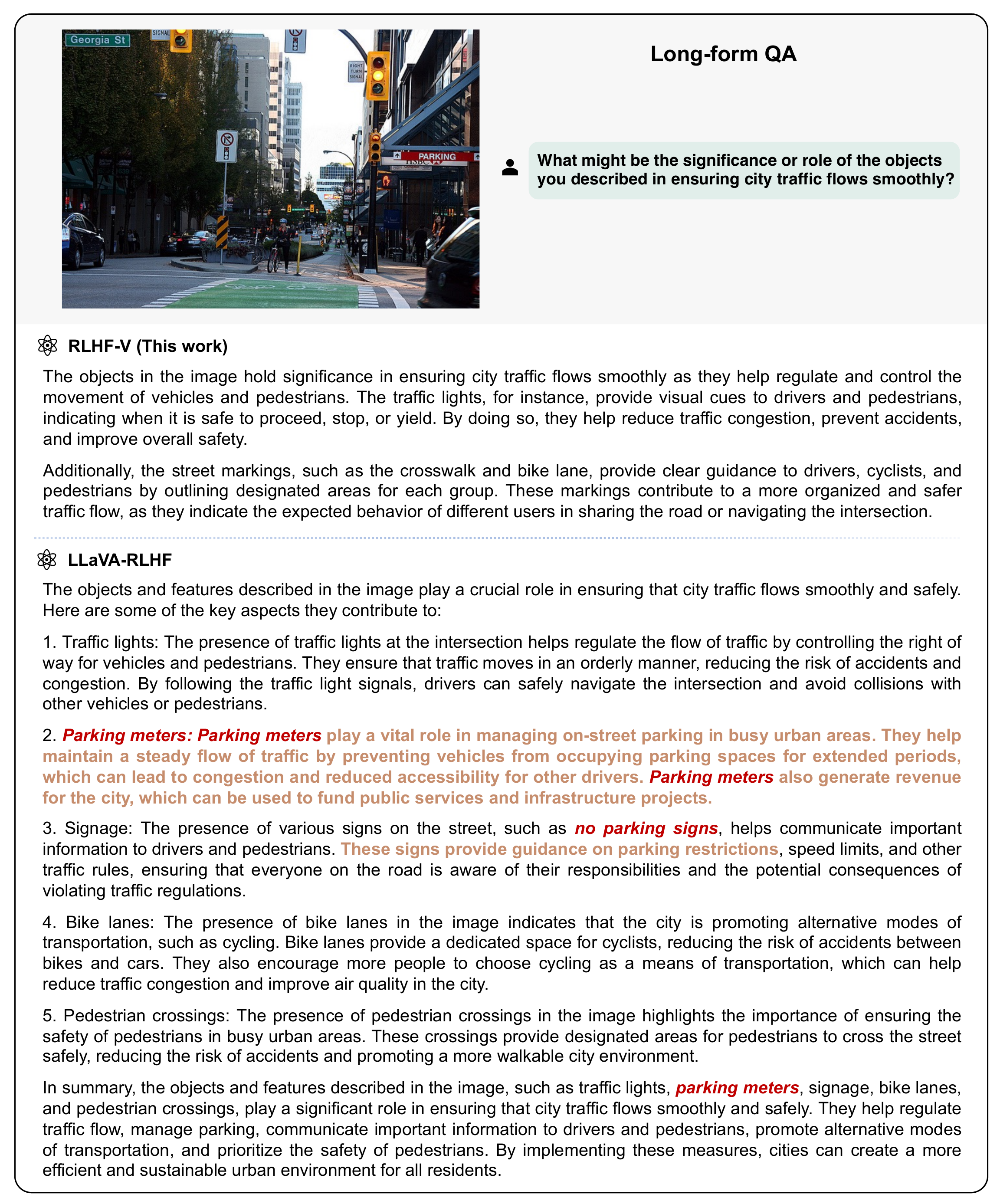}
    
    \caption{Qualitative results of different models. 
 {\color{yellow}\textbf{Unreasonable reasoning and extensions}} and {\color{darkred}\textbf{\textit{scene related hallucinations}}} are highlighted in color respectively.}
    \label{fig:case_5}
\end{figure*}

\begin{figure*}
    \centering

    \includegraphics[width=0.95\linewidth]{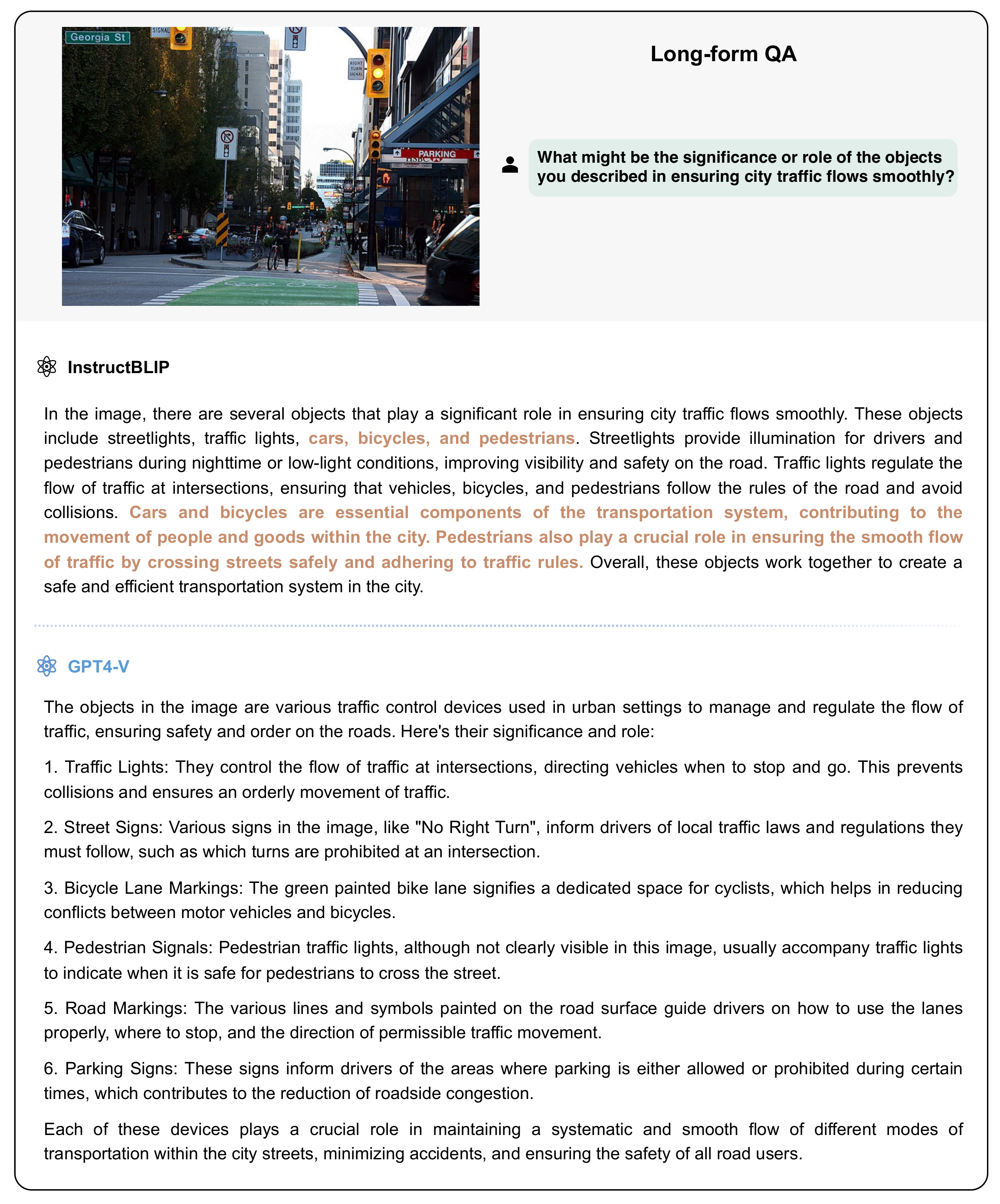}
    
    \caption{Qualitative results of different models (continued figure). 
 {\color{yellow}\textbf{Unreasonable reasoning and extensions}} are highlighted in color.}
    \label{fig:case_5-2}
\end{figure*}

\begin{figure*}
    \centering

    \includegraphics[width=0.95\linewidth]{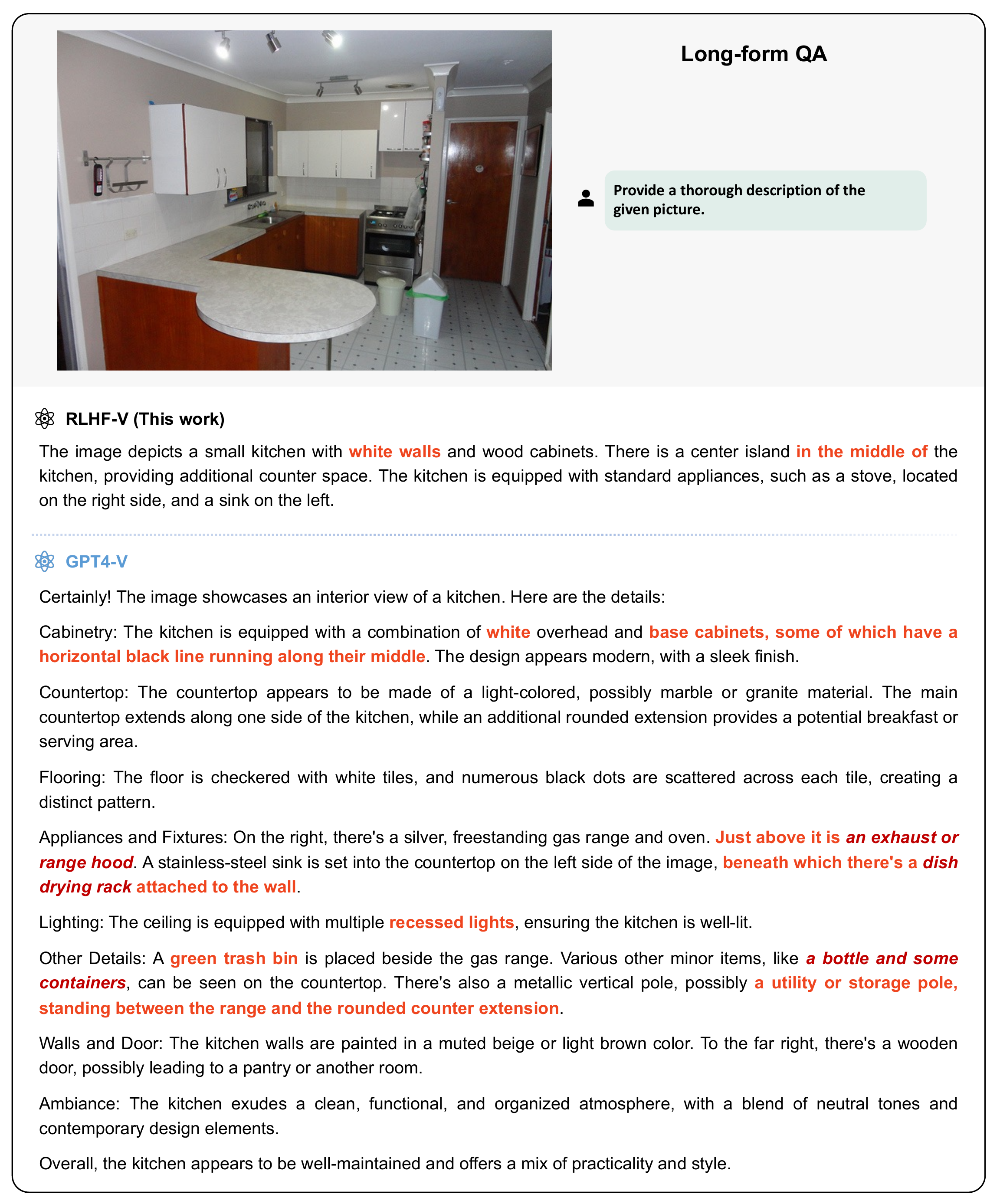}
    
    \caption{Qualitative results of different models. 
   {\color{red}\textbf{Hallucinations}} and {\color{darkred}\textbf{\textit{scene related hallucinations}}} are highlighted in color respectively.}
    \label{fig:case_3}
\end{figure*}

\begin{figure*}
    \centering

    \includegraphics[width=0.95\linewidth]{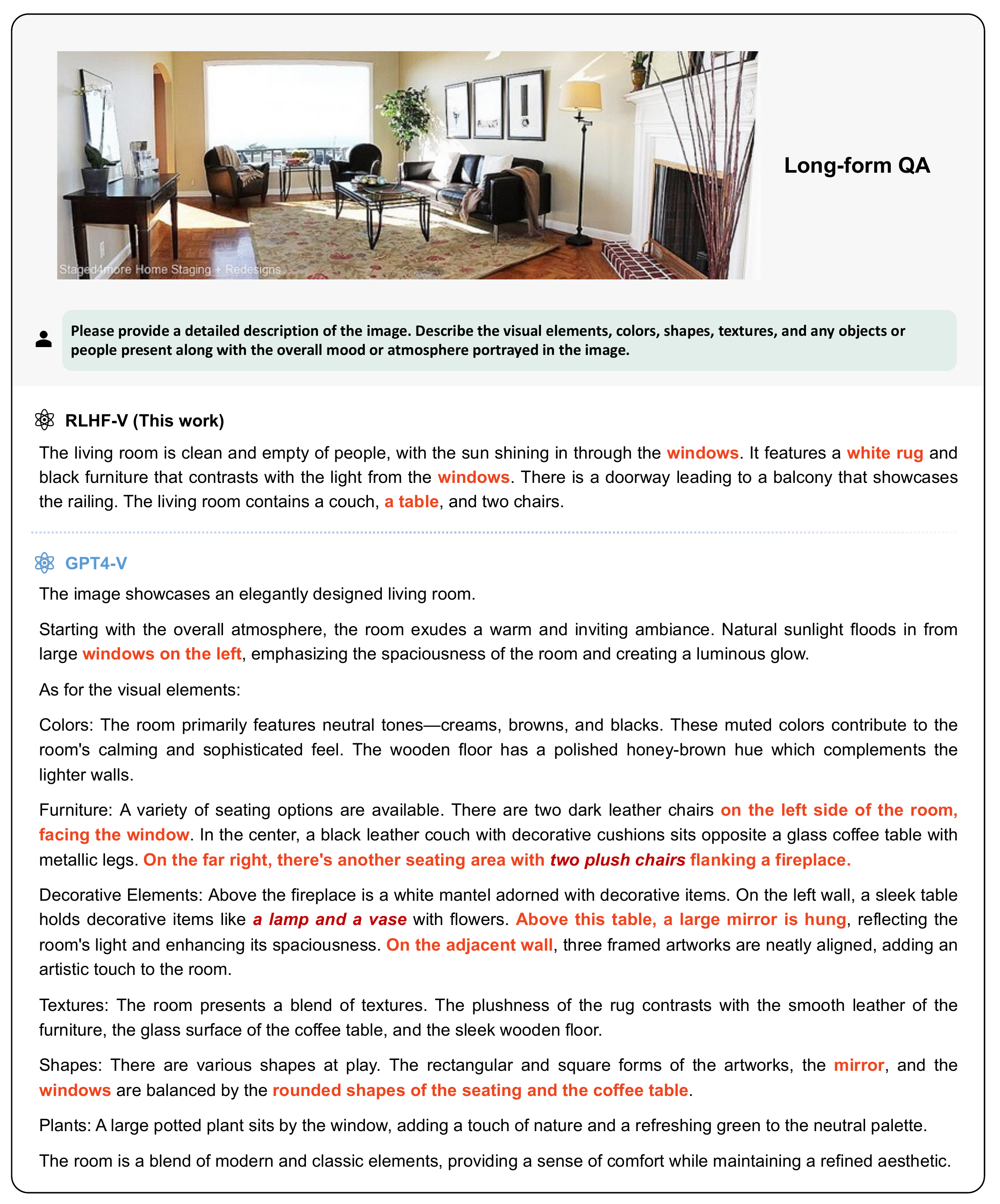}
    
    \caption{Qualitative results of different models. 
 {\color{red}\textbf{Hallucinations}} and {\color{darkred}\textbf{\textit{scene related hallucinations}}} are highlighted in color respectively.}
    \label{fig:case_4}
\end{figure*}

% WARNING: do not forget to delete the supplementary pages from your submission 
% \input{sec/X_suppl}

\end{document}